\documentclass[11pt]{article}
\usepackage{acl2016}
\usepackage{times}
\usepackage{url}
\usepackage{latexsym}

\usepackage{graphicx}
\usepackage{booktabs}
\usepackage{tabularx}

\aclfinalcopy 

\title{Compositional Sequence Labeling Models for Error Detection \\ in Learner Writing}

\author{\hspace{-0.5cm}Marek Rei\\
	    \hspace{-0.5cm}The ALTA Institute\\
	    \hspace{-0.5cm}Computer Laboratory\\
	    \hspace{-0.5cm}University of Cambridge\\
        \hspace{-0.5cm}United Kingdom\\
	    \hspace{-0.5cm}{\tt marek.rei@cl.cam.ac.uk}
	    \And
	    \hspace{0.5cm}Helen Yannakoudakis\\
	    \hspace{0.5cm}The ALTA Institute\\
	    \hspace{0.5cm}Computer Laboratory\\
	    \hspace{0.5cm}University of Cambridge\\
        \hspace{0.5cm}United Kingdom\\
        \hspace{0.5cm}{\tt helen.yannakoudakis@cl.cam.ac.uk}}

\date{}

\begin{document}
\maketitle
\begin{abstract}
In this paper, we present the first experiments using neural network models for the task of error detection in learner writing.
We perform a systematic comparison of alternative compositional architectures and propose a framework for error detection based on bidirectional LSTMs.
Experiments on the CoNLL-14 shared task dataset show the model is able to outperform other participants on detecting errors in learner writing.
Finally, the model is integrated with a publicly deployed self-assessment system, leading to performance comparable to human annotators.

\end{abstract}

\section{Introduction}

%


Automated systems for detecting errors in learner writing are valuable tools for second language learning and assessment.
Most work in recent years has focussed on error correction, with error detection performance measured as a byproduct of the correction output \cite{Ng2013,Ng2013a}. 
However, this assumes that systems are able to propose a correction for every detected error, and accurate systems for correction might not be optimal for detection.
While  closed-class errors such as incorrect prepositions and determiners can be modeled with a supervised classification approach, content-content word errors are the 3rd most frequent error type and pose a serious challenge to error correction frameworks \cite{Leacock2014,Kochmar2014}.
Evaluation of error correction is also highly subjective and human annotators have rather low agreement on gold-standard corrections \cite{Bryant2015}.
Therefore, we treat error detection in learner writing as an independent task and propose a system for labeling each token as being correct or incorrect in context.



Common approaches to similar sequence labeling tasks involve learning weights or probabilities for context n-grams of varying sizes, or relying on previously extracted high-confidence context patterns. Both of these methods can suffer from data sparsity, as they treat words as independent units and miss out on potentially related patterns. In addition, they need to specify a fixed context size and are therefore often limited to using a small window near the target.

Neural network models aim to address these weaknesses and have achieved success in various NLP tasks such as language modeling \cite{Bengio2003} and speech recognition \cite{Dahl2012}.
Recent developments in machine translation have also shown that text of varying length can be represented as a fixed-size vector using convolutional networks \cite{Kalchbrenner2013,Cho2014} or recurrent neural networks \cite{Cho2014a,Bahdanau2015}. 

In this paper, we present the first experiments using neural network models for the task of error detection in learner writing.
We perform a systematic comparison of alternative compositional structures for constructing informative context representations.
Based on the findings, we propose a novel framework for performing error detection in learner writing, which achieves state-of-the-art results on two datasets of error-annotated learner essays.
The sequence labeling model creates a single variable-size network over the whole sentence, conditions each label on all the words, and predicts all labels together.
The effects of different datasets on the overall performance are investigated by incrementally providing additional training data to the model.
Finally, we integrate the error detection framework with a publicly deployed self-assessment system, leading to performance comparable to human annotators.

\section{Background and Related Work}

The field of automatically detecting errors in learner text has a long and rich history.
Most work has focussed on tackling specific types of errors, such as usage of incorrect prepositions \cite{Tetreault2008,Chodorow2007}, articles \cite{Han2004,Han2006}, verb forms \cite{Lee2008}, and adjective-noun pairs \cite{Kochmar2014}. 

However, there has been limited work on more general error detection systems that could handle all types of errors in learner text.
\newcite{Chodorow1998} proposed a method based on mutual information and the chi-square statistic to detect sequences of part-of-speech tags and function words that are likely to be ungrammatical in English.
\newcite{Gamon2011} used Maximum Entropy Markov Models with a range of features, such as POS tags, string features, and outputs from a constituency parser. 
The pilot Helping Our Own shared task \cite{Dale2011} also evaluated grammatical error detection of a number of different error types, though most systems were error-type specific and the best approach was heavily skewed towards article and preposition errors \cite{Rozovskaya2011}.  
We extend this line of research, working towards general error detection systems, and investigate the use of neural compositional models on this task.

The related area of grammatical error \textit{correction} has also gained considerable momentum in the past years, with four recent shared tasks highlighting several emerging directions \cite{Dale2011,Dale2012,Ng2013,Ng2013a}. 
The current state-of-the-art approaches can broadly be separated into two categories:

\begin{enumerate}
\item Phrase-based statistical machine translation techniques, essentially translating the incorrect source text into the corrected version \cite{Felice2014,Junczys-Dowmunt2014}
\item Averaged Perceptrons and Naive Bayes classifiers making use of native-language error correction priors \cite{Rozovskaya2014,Rozovskaya2013}.
\end{enumerate}

\noindent Error correction systems require very specialised models, as they need to generate an improved version of the input text, whereas a wider range of tagging and classification models can be deployed on error detection.
In addition, automated writing feedback systems that indicate the presence and location of errors may be better from a pedagogic point of view, rather than providing a panacea and correcting all errors in learner text.
In Section \ref{sec:conll} we evaluate a neural sequence tagging model on the latest shared task test data, and compare it to the top participating systems on the task of error detection.

\section{Sequence Labeling Architectures}

\begin{figure*}[t]
	\includegraphics[width=\linewidth]{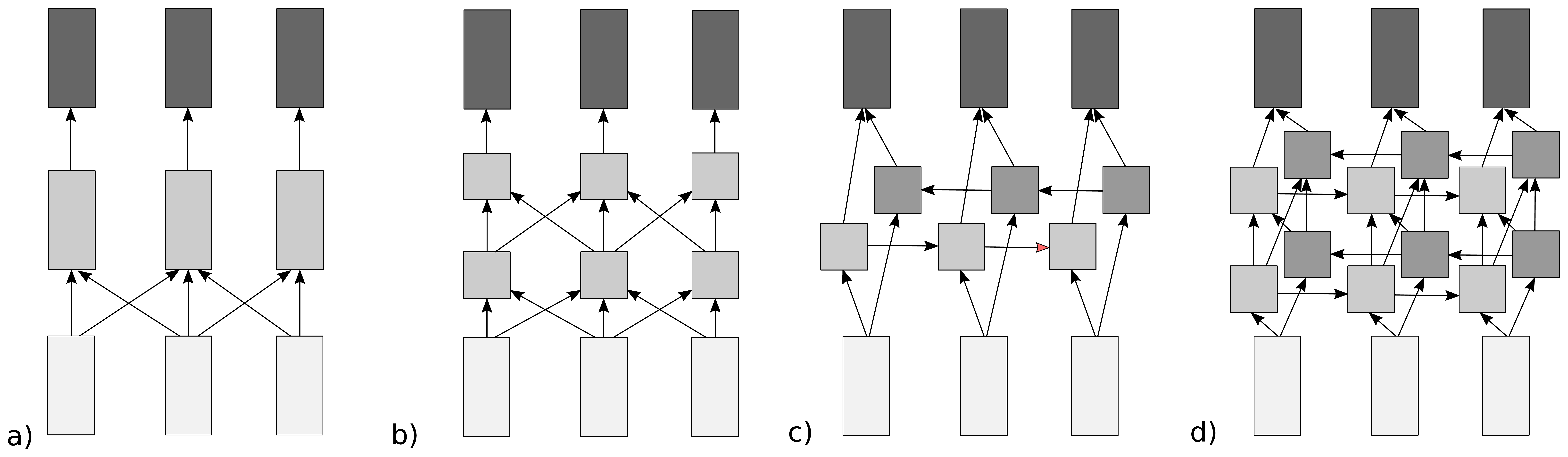}
	\caption{Alternative neural composition architectures for error detection. a) Convolutional network b) Deep convolutional network c) Recurrent bidirectional network d) Deep recurrent bidirectional network. The bottom layers are embeddings for individual tokens. The middle layers are context-dependent representations, built using different composition functions. The top layers are softmax output layers, predicting a label distribution for every input token.}
	\label{fig:networks}
\end{figure*}

We construct a neural network sequence labeling framework for the task of error detection in learner writing.
The model receives only a series of tokens as input, and outputs the probability of each token in the sentence being correct or incorrect in a given context.
The architectures start with the vector representations of individual words, $[x_1, ... , x_T]$, where $T$ is the length of the sentence. 
Different composition functions are then used to calculate a hidden vector representation of each token in context, $[h_1, ..., h_T]$.
These representations are passed through a softmax layer, producing a probability distribution over the possible labels for every token in context:

\begin{equation}
p_t = softmax(W_o h_t)
\end{equation}

\noindent where $W_o$ is the weight matrix between the hidden vector $h_t$ and the output layer. 

We investigate six alternative neural network architectures for the task of error detection: convolutional, bidirectional recurrent, bidirectional LSTM, and multi-layer variants of each of them.
In the \textbf{convolutional} neural network (CNN, Figure \ref{fig:networks}a) for token labeling, the hidden vector $h_t$ is calculated based on a fixed-size context window.
The convolution acts as a feedforward network, using surrounding context words as input, and therefore it will learn to detect the presence of different types of n-grams.
The assumption behind the convolutional architecture is that memorising erroneous token sequences from the training data is sufficient for performing error detection. 

The convolution uses $d_w$ tokens on either side of the target token, and the vectors for these tokens are concatenated, preserving the ordering:

\begin{equation}
c_t = x_{t-d_w} : ... : x_{t+d_w}
\end{equation}

\noindent where $x_1:x_2$ is used as notation for vector concatenation of $x_1$ and $x_2$.
The combined vector is then passed through a non-linear layer to produce the hidden representation:

\begin{equation}
h_t = tanh(W_c  c_t)
\end{equation}

The \textbf{deep convolutional} network (Figure \ref{fig:networks}b) adds an extra convolutional layer to the architecture, using the first layer as input. It creates convolutions of convolutions, thereby capturing more complex higher-order features from the dataset.

In a \textbf{recurrent} neural network (RNN), each hidden representation is calculated based on the current token embedding and the hidden vector at the previous time step:

\begin{equation}
h_t = f(W x_t + V h_{t-1})
\end{equation} 

\noindent where $f(z)$ is a nonlinear function, such as the sigmoid function. 
Instead of a fixed context window, information is passed through the sentence using a recursive function and the network is able to learn which patterns to disregard or pass forward.
This recurrent network structure is referred to as an Elman-type network, after \newcite{Elman1990}.

The \textbf{bidirectional RNN} (Figure \ref{fig:networks}c) consists of two recurrent components, moving in opposite directions through the sentence. While the unidirectional version takes into account only context on the left of the target token, the bidirectional version recursively builds separate context representations from either side of the target token. The left and right context are then concatenated and used as the hidden representation:

\begin{equation}
h_t^{\rightarrow} = f(W_r x_t + V_r h_{t-1}^{\rightarrow})
\end{equation} 

\begin{equation}
h_t^{\leftarrow} = f(W_l x_t + V_l h_{t+1}^{\leftarrow})
\end{equation} 

\begin{equation}
h_t = h_t^{\rightarrow} : h_t^{\leftarrow}
\end{equation}

Recurrent networks have been shown to perform well on the task of language modeling \cite{Kombrinka,Chelba2014}, where they learn an incremental composition function for predicting the next token in the sequence. 
However, while language models can estimate the probability of each token, they are unable to differentiate between infrequent and incorrect token sequences.
For error detection, the composition function needs to learn to identify semantic anomalies or ungrammatical combinations, independent of their frequency.
The bidirectional model provides extra information, as it allows the network to use context on both sides of the target token.

\newcite{Irsoy2014a} created an extension of this architecture by connecting together multiple layers of bidirectional Elman-type recurrent network modules. This \textbf{deep bidirectional RNN} (Figure \ref{fig:networks}d) calculates a context-dependent representation for each token using a bidirectional RNN, and then uses this as input to another bidirectional RNN.
The multi-layer structure allows the model to learn more complex higher-level features and effectively perform multiple recurrent passes through the sentence.

The long-short term memory (\textbf{LSTM}) \cite{Hochreiter1997} is an advanced alternative to the Elman-type networks that has recently become increasingly popular. 
It uses two separate hidden vectors to pass information between different time steps, and includes gating mechanisms for modulating its own output.
LSTMs have been successfully applied to various tasks, such as speech recognition \cite{Graves2013a}, machine translation \cite{Luong2015}, and natural language generation \cite{Wen2015}. 

Two sets of gating values (referred to as the \textit{input} and \textit{forget} gates) are first calculated based on the previous states of the network:
\begin{equation}
i_t = \sigma(W_i x_t + U_i h_{t-1} + V_f c_{t-1} + b_i)
\end{equation}
\begin{equation}
f_t = \sigma(W_f x_t + U_f h_{t-1} + V_f c_{t-1} + b_f)
\end{equation}

\noindent where $x_t$ is the current input, $h_{t-1}$ is the previous hidden state, $b_i$ and $b_f$ are biases, $c_{t-1}$ is the previous internal state (referred to as the \textit{cell}), and $\sigma$ is the logistic function.
The new internal state is calculated based on the current input and the previous hidden state, and then interpolated with the previous internal state using $f_t$ and $i_t$
 as weights:
\begin{equation}
\widetilde{c_t} = tanh(W_c x_t + U_c h_{t-1} + b_c)
\end{equation}
\begin{equation}
\label{eq:lstm_interpolation}
c_t = f_t \odot c_{t-1} + i_t \odot \widetilde{c_t}
\end{equation}

\noindent where $\odot$ is element-wise multiplication.
Finally, the hidden state is calculated by passing the internal state through a $tanh$ nonlinearity, and weighting it with $o_t$. The values of $o_t$ are conditioned on the new internal state ($c_{t}$), as opposed to the previous one ($c_{t-1}$):
\begin{equation}
o_t = \sigma(W_o x_t + U_o h_{t-1} + V_o c_{t} + b_o)
\end{equation}
\begin{equation}
h_t = o_t \odot tanh(c_t)
\end{equation}

Because of the linear combination in equation (\ref{eq:lstm_interpolation}), the LSTM is less susceptible to vanishing gradients over time, thereby being able to make use of longer context when making predictions. In addition, the network learns to modulate itself, effectively using the gates to predict which operation is required at each time step, thereby incorporating higher-level features.

In order to use this architecture for error detection, we create a \textbf{bidirectional LSTM}, making use of the advanced features of LSTM and incorporating context on both sides of the target token. In addition, we experiment with a \textbf{deep bidirectional LSTM}, which includes two consecutive layers of bidirectional LSTMs, modeling even more complex features and performing multiple passes through the sentence.

\begin{table*}
\setlength\tabcolsep{11.0pt}
\begin{tabular}{lrrr|rrrrr} \toprule
 & \multicolumn{3}{c|}{Development}  & \multicolumn{5}{c}{Test} \\
 & P & R & $F_{0.5}$ & predicted & correct & P & R & $F_{0.5}$ \\ \midrule
CRF & 62.2 & 13.6 & 36.3 & 914 & 516 & 56.5 & 8.2 & 25.9 \\ \midrule
CNN & 52.4 & 24.9 & 42.9 & 3518 & 1620 & 46.0 & 25.7 & 39.8 \\
Deep CNN & 48.4 & 26.2 & 41.4 & 3992 & 1651 & 41.4 & 26.2 & 37.1 \\
Bi-RNN & \textbf{63.9} & 18.0 & 42.3 & 2333 & 1196 & \textbf{51.3} & 19.0 & 38.2 \\
Deep Bi-RNN & 60.3 & 17.6 & 40.6 & 2543 & 1255 & 49.4 & 19.9 & 38.1 \\
Bi-LSTM & 54.5 & \textbf{28.2} & \textbf{46.0} & 3898 & 1798 & 46.1 & \textbf{28.5} & \textbf{41.1} \\
Deep Bi-LSTM & 56.7 & 21.3 & 42.5 & 2822 & 1359 & 48.2 & 21.6 & 38.6 \\ \bottomrule
\end{tabular}
\caption{Performance of the CRF and alternative neural network structures on the public FCE dataset for token-level error detection in learner writing.}
\label{tab:structure}
\end{table*}

For comparison with non-neural models, we also report results using \textbf{CRFs} \cite{Lafferty2001}, which are a popular choice for sequence labeling tasks. We trained the CRF++ \footnote{https://taku910.github.io/crfpp/} implementation on the same dataset, using as features unigrams, bigrams and trigrams in a 7-word window surrouding the target word (3 words before and after). The predicted label is also conditioned on the previous label in the sequence.

\section{Experiments}
\label{sec:experiments}

We evaluate the alternative network structures on the publicly released First Certificate in English dataset  (FCE-public, \newcite{Yannakoudakis2011}). The dataset contains short texts, written by learners of English as an additional language in response to exam prompts eliciting free-text answers and assessing mastery of the upper-intermediate proficiency level. 
The texts have been manually error-annotated using a taxonomy of 77 error types. 
We use the released test set for evaluation, containing 2,720 sentences, leaving 30,953 sentences for training.
We further separate 2,222 sentences from the training set for development and hyper-parameter tuning.

The dataset contains manually annotated error spans of various types of errors, together with their suggested corrections. We convert this to a token-level error detection task by labeling each token inside the error span as being incorrect. 
In order to capture errors involving missing words, the error label is assigned to the token immediately after the incorrect gap -- this is motivated by the intuition that while this token is correct when considered in isolation, it is incorrect in the current context, as another token should have preceeded it.

As the main evaluation measure for error detection we use $F_{0.5}$, which was also the measure adopted in the CoNLL-14 shared task on error correction \cite{Ng2013a}. It combines both precision and recall, while assigning twice as much weight to precision, since accurate feedback is often more important than coverage in error detection applications \cite{Nagata2010}.
Following \newcite{Chodorow2012}, we also report raw counts for predicted and correct tokens.
Related evaluation measures, such as the $M^2$-scorer \cite{Ng2013a} and the I-measure \cite{Felice2015}, require the system to propose a correction and are therefore not directly applicable on the task of error detection.

During the experiments, the input text was lowercased and all tokens that occurred less than twice in the training data were represented as a single \textit{unk} token.
Word embeddings were set to size $300$ and initialised using the publicly released pretrained Word2Vec vectors \cite{Mikolov2013a}.
The convolutional networks use window size $3$ on either side of the target token and produce a 300-dimensional context-dependent vector.
The recurrent networks use hidden layers of size 200 in either direction.
We also added an extra hidden layer of size $50$ between each of the composition functions and the output layer -- this allows the network to learn a separate non-linear transformation and reduces the dimensionality of the compositional vectors.
The parameters were optimised using gradient descent with initial learning rate $0.001$, the {\small ADAM} algorithm \cite{Kingma2015} for dynamically adapting the learning rate, and batch size of 64 sentences.
$F_{0.5}$ on the development set was evaluated at each epoch, and the best model was used for final evaluations.

\section{Results}

Table \ref{tab:structure} contains results for experiments comparing different composition architectures on the task of error detection.
The CRF has the lowest $F_{0.5}$ score compared to any of the neural models. It memorises frequent error sequences with high precision, but does not generalise sufficiently, resulting in low recall.
The ability to condition on the previous label also does not provide much help on this task -- there are only two possible labels and the errors are relatively sparse.

The architecture using convolutional networks performs well and achieves the second-highest result on the test set. It is designed to detect error patterns from a fixed window of 7 words, which is large enough to not require the use of more advanced composition functions.
In contrast, the performance of the bidirectional recurrent network (Bi-RNN) is somewhat lower, especially on the test set.
In Elman-type recurrent networks, the context signal from distant words decreases fairly rapidly due to the sigmoid activation function and diminishing gradients. 
This is likely why the Bi-RNN achieves the highest precision of all systems -- the predicted label is mostly influenced by the target token and its immediate neighbours, allowing the network to only detect short high-confidence error patterns.
The convolutional network, which uses 7 context words with equal attention, is able to outperform the Bi-RNN despite the fixed-size context window.

The best overall result and highest $F_{0.5}$ is achieved by the bidirectional LSTM composition model (Bi-LSTM). 
This architecture makes use of the full sentence for building context vectors on both sides of the target token, but improves on Bi-RNN by utilising a more advanced composition function. Through the application of a linear update for the internal cell representation, the LSTM is able to capture dependencies over longer distances. In addition, the gating functions allow it to adaptively decide which information to include in the hidden representations or output for error detection.

We found that using multiple layers of compositional functions in a deeper network gave comparable or slightly lower results for all the composition architectures. 
This is in contrast to \newcite{Irsoy2014a}, who experimented with Elman-type networks and found some improvements using multiple layers of Bi-RNNs.
The differences can be explained by their task benefiting from alternative features: the evaluation was performed on opinion mining where most target sequences are longer phrases that need to be identified based on their semantics, whereas many errors in learner writing are short and can only be identified by a contextual mismatch. In addition,
our networks contain an extra hidden layer before the output, which allows the models to learn higher-level representations without adding complexity through an extra compositional layer.

\section{Additional Training Data}
\label{sec:bigdata}

\begin{table}[t]
\setlength\tabcolsep{14.5pt}
\begin{tabular}{lrr} \toprule
Training data & Dev $F_{0.5}$ & Test $F_{0.5}$ \\ \midrule
FCE-public & 46.0 & 41.1 \\
+NUCLE & 39.0 & 41.0 \\
+IELTS & 45.6 & 50.7 \\
+FCE & 57.2 & 61.1 \\
+CPE & 59.0 & 62.1 \\
+CAE & \textbf{60.7} & \textbf{64.3} \\ \bottomrule
\end{tabular}
\caption{Results on the public FCE test set when incrementally providing more training data to the error detection model.}
\label{tab:bigdata}
\end{table}

There are essentially infinitely many ways of committing errors in text and introducing additional training data should alleviate some of the problems with data sparsity. We experimented with incrementally adding different error-tagged corpora into the training set and measured the resulting performance. This allows us to provide some context to the results obtained by using each of the datasets, and gives us an estimate of how much annotated data is required for optimal performance on error detection.
The datasets we consider are as follows:

\begin{itemize}
\setlength\itemsep{-0.1em}
\item FCE-public -- the publicly released subset of FCE \cite{Yannakoudakis2011}, as described in Section \ref{sec:experiments}.
\item NUCLE -- the NUS Corpus of Learner English \cite{Dahlmeier2013}, used as the main training set for CoNLL shared tasks on error correction.
\item IELTS -- a subset of the IELTS examination dataset extracted from the Cambridge Learner Corpus (CLC, \newcite{Nicholls2003}), containing 68,505 sentences from all proficiency levels, also used by \newcite{Felice2014}.
\item FCE -- a larger selection of FCE texts from the CLC, containing 323,192 sentences.
\item CPE -- essays from the proficient examination level in the CLC, containing 210,678 sentences.
\item CAE -- essays from the advanced examination level in the CLC, containing 219,953 sentences.

\end{itemize}

Table \ref{tab:bigdata} contains results obtained by incrementally adding training data to the Bi-LSTM model.
We found that incorporating the NUCLE dataset does not improve performance over using only the FCE-public dataset, which is likely due to the two corpora containing texts with different domains and writing styles.
The texts in FCE are written by young intermediate students, in response to prompts eliciting letters, emails and reviews, whereas NUCLE contains mostly argumentative essays written by advanced adult learners.
The differences in the datasets offset the benefits from additional training data, and the performance remains roughly the same.

\begin{figure}[h]
	\includegraphics[width=\linewidth]{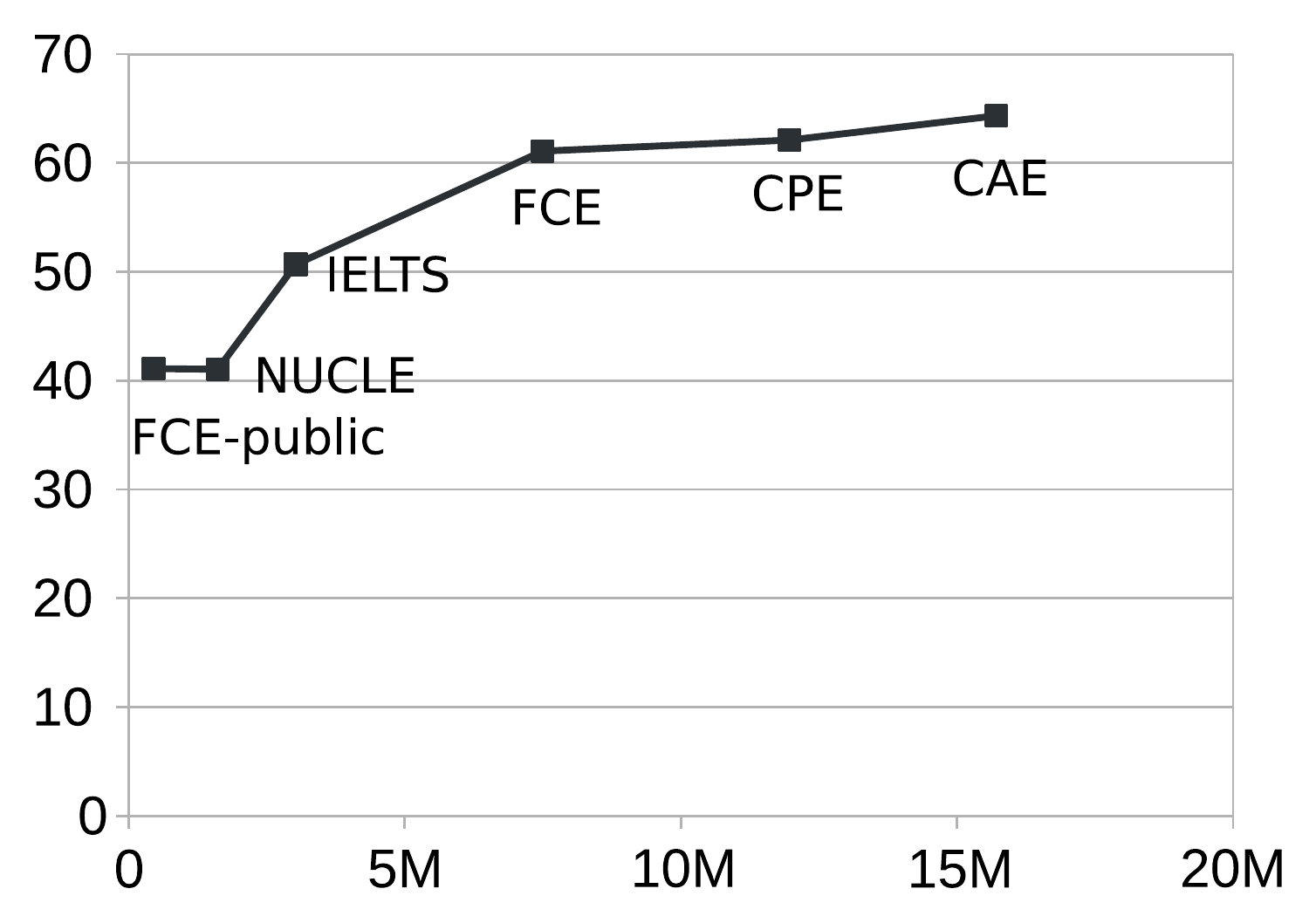}
	\caption{$F_{0.5}$ measure on the public FCE test set, as a function of the total number of tokens in the training set.}
	\label{fig:bigdata}
\end{figure}

In contrast, substantial improvements are obtained when introducing the IELTS and FCE datasets, with each of them increasing the $F_{0.5}$ score by roughly $10\%$.
The IELTS dataset contains essays from all proficiency levels, and FCE from mid-level English learners, which provides the model with a distribution of `average' errors to learn from. Adding even more training data from high-proficiency essays in CPE and CAE only provides minor further improvements.

\begin{table*}
\begin{tabular}{lr|rrrr|rrrr} \toprule
 &  & \multicolumn{4}{c|}{Annotation 1} & \multicolumn{4}{c}{Annotation 2} \\ \midrule
 & predicted & correct & P & R & $F_{0.5}$ & correct & P & R & $F_{0.5}$ \\ \midrule
Annotator 1 & 2992 & - & - & - & - & 1800 & 60.2 & 42.9 & 55.7 \\
Annotator 2 & 4199 & 1800 & 42.9 & 60.2 & 45.5 & - & - & - & - \\ \midrule
CAMB & 2170 & 731 & 33.7 & \textbf{24.4} & 31.3 & 1052 & 48.5 & \textbf{25.1} & 40.8 \\
CUUI & 1582 & 550 & 34.8 & 18.4 & 29.5 & 755 & 47.7 & 18.0 & 35.9 \\
AMU  & 1260 & 479 & 38.0 & 16.0 & 29.8 & 643 & 51.0 & 15.3 & 34.8 \\
P1+P2+S1+S2 & 887 & 388 & \textbf{43.7} & 13.0 & 29.7 & 535 & \textbf{60.3} & 12.7 & 34.5 \\ \midrule
Bi-LSTM (FCE-public) & 4449 & 683 & 15.4 & 22.8 & 16.4 & 1052 & 23.6 & 25.1 & 23.9  \\
Bi-LSTM (full) & 1540 & 627 & 40.7 & 21.0 & \textbf{34.3} & 911 & 59.2 & 21.7 & \textbf{44.0} \\ \bottomrule
\end{tabular}
\caption{Error detection results on the two official annotations for the CoNLL-14 shared task test dataset.}
\label{tab:conll}
\end{table*}

Figure \ref{fig:bigdata} also shows $F_{0.5}$ on the FCE-public test set as a function of the total number of tokens in the training data. The optimal trade-off between performance and data size is obtained at around 8 million tokens, after introducing the FCE dataset.

\section{CoNLL-14 Shared Task}
\label{sec:conll}


The CoNLL-14 shared task \cite{Ng2013a} focussed on automatically correcting errors in learner writing. The NUCLE dataset was provided as the main training dataset, but participants were allowed to include other annotated corpora and external resources. For evaluation, 25 students were recruited to each write two new essays, which were then annotated by two experts.

We used the same methods from Section \ref{sec:experiments} for converting the shared task annotation to a token-level labeling task in order to evaluate the models on error detection. In addition, the correction outputs of all the participating systems were made available online, therefore we are able to report their performance on this task. 
In order to convert their output to error detection labels, the corrected sentences were aligned with the original input using Levenshtein distance, and any changes proposed by the system resulted in the corresponding source words being labeled as errors.

The results on the two annotations of the shared task test data can be seen in Table \ref{tab:conll}.
We first evaluated each of the human annotators with respect to the other, in order to estimate the upper bound on this task. The average $F_{0.5}$ of roughly 50\% shows that the task is difficult and even human experts have a rather low agreement. It has been shown before that correcting grammatical errors is highly subjective \cite{Bryant2015}, but these results indicate that trained annotators can disagree even on the number and location of errors.

In the same table, we provide error detection results for the top 3 participants in the shared task: CAMB \cite{Felice2014}, CUUI \cite{Rozovskaya2014}, and AMU \cite{Junczys-Dowmunt2014}. They each preserve their relative ranking also in the error detection evaluation. The CAMB system has a lower precision but the highest recall, also resulting in the highest $F_{0.5}$. CUUI and AMU are close in performance, with AMU having slightly higher precision. 

After the official shared task, \newcite{Susanto2014} published a system which combines several alternative models and outperforms the shared task participants when evaluated on error correction. However, on error detection it receives lower results, ranking 3rd and 4th when evaluated on $F_{0.5}$  (P1+P2+S1+S2 in Table \ref{tab:conll}). The system has detected a small number of errors with high precision, and does not reach the highest $F_{0.5}$.

Finally, we present results for the Bi-LSTM sequence labeling system for error detection.
Using only FCE-public for training, the overall performance is rather low as the training set is very small and contains texts from a different domain. 
However, these results show that the model behaves as expected -- since it has not seen similar language during training, it labels a very large portion of tokens as errors.
This indicates that the network is trying to learn correct language constructions from the limited data and classifies unseen structures as errors, as opposed to simply memorising error sequences from the training data.

When trained on all the datasets from Section \ref{sec:bigdata}, the model achieves the highest $F_{0.5}$ of all systems on both of the CoNLL-14 shared task test annotations, with an absolute improvement of $3\%$ over the previous best result. 
It is worth noting that the full Bi-LSTM has been trained on more data than the other CoNLL contestants. However, as the shared task systems were not restricted to the NUCLE training set, all the submissions also used differing amounts of training data from various sources.
In addition, the CoNLL systems are mostly combinations of many alternative models: the CAMB system is a hybrid of machine translation, a rule-based system, and a language model re-ranker; CUUI consists of different classifiers for each individual error type; and P1+P2+S1+S2 is a combination of four different error correction systems.
In contrast, the Bi-LSTM is a single model for detecting all error types, and therefore represents a more scalable data-driven approach.

\section{Essay Scoring}

In this section, we perform an extrinsic evaluation of the efficacy of the error detection system and examine the extent to which it generalises at higher levels of granularity on the task of automated essay scoring. More specifically, we replicate experiments using the \textit{text-level} model described by \newcite{Andersen2013}, which is currently deployed in a self-assessment and tutoring system (SAT), an online automated writing feedback tool actively used by language learners.\footnote{http://www.cambridgeenglish.org/learning-english/free-resources/write-and-improve/}

The SAT system predicts an overall score for a given text, which provides a holistic assessment of linguistic competence and language proficiency. The authors trained a supervised ranking perceptron model on the FCE-public dataset, using features such as error-rate estimates from a language model and various lexical and grammatical properties of text (e.g., word n-grams, part-of-speech n-grams and phrase-structure rules).
We replicate this experiment and add the average probability of each token in the essay being correct, according to the error detection model, as an additional feature for the scoring framework.
The system was then retrained on FCE-public and evaluated on correctly predicting the assigned essay score. 
Table \ref{tab:results_script} presents the experimental results. 

\begin{table}[h]
\centering
\begin{tabular}{lcc} \toprule
 & $r$ & $\rho$ \\ \midrule
Human annotators & 79.6 & 79.2 \\
SAT & 75.1 & 76.0 \\
SAT + Bi-LSTM (FCE-public) & 76.0 & 77.0 \\
SAT + Bi-LSTM (full) & \textbf{78.0} & \textbf{79.9} \\ \bottomrule
\end{tabular}
\caption{Pearson's correlation $r$  and Spearman's correlation $\rho$ on the public FCE test set on the task of automated essay scoring.}
\label{tab:results_script}
\end{table}

The human performance on the test set is calculated as the average inter-annotator correlation on the same data, and the existing SAT system has demonstrated levels of performance that are very close to that of human assessors.
Nevertheless, the Bi-LSTM model trained only on FCE-public complements the existing features, and the combined model achieves an absolute improvement of around 1\% percent, corresponding to 20-31\% relative error reduction with respect to the human performance.
Even though the Bi-LSTM is trained on the same dataset and the SAT system already includes various linguistic features for capturing errors, our error detection model manages to further improve its performance.

When the Bi-LSTM is trained on all the available data from Section \ref{sec:bigdata}, the combination achieves further substantial improvements. The relative error reduction on Pearson's correlation is 64\%, and the system actually outperforms human annotators on Spearman's correlation.



\section{Conclusions}

In this paper, we presented the first experiments using neural network models for the task of error detection in learner writing.
Six alternative compositional network architectures for modeling context were evaluated.
Based on the findings, we propose a novel error detection framework using token-level embeddings, bidirectional LSTMs for context representation, and a multi-layer architecture for learning more complex features.
This structure allows the model to classify each token as being correct or incorrect, using the full sentence as context. The self-modulation architecture of LSTMs was also shown to be beneficial, as it allows the network to learn more advanced composition rules and remember dependencies over longer distances.

Substantial performance improvements were achieved by training the best model on additional datasets. We found that the largest benefit was obtained from training on 8 million tokens of text from learners with varying levels of language proficiency. In contrast, including even more data from higher-proficiency learners gave marginal further improvements.
As part of future work, it would be beneficial to investigate the effect of automatically generated training data for error detection (e.g., \newcite{Rozovskaya2010}).

We evaluated the performance of existing error correction systems from CoNLL-14 on the task of error detection.
The experiments showed that success on error correction does not necessarily mean success on error detection, as the current best correction system (P1+P2+S1+S2) is not the same as the best shared task detection system (CAMB).
In addition, the neural sequence tagging model, specialised for error detection, was able to outperform all other participating systems.

Finally, we performed an extrinsic evaluation by incorporating probabilities from the error detection system as features in an essay scoring model. Even without any additional data, the combination further improved performance which is already close to the results from human annotators.
In addition, when the error detection model was trained on a larger training set, the essay scorer was able to exceed human-level performance.


\section*{Acknowledgments}

We would like to thank Prof Ted Briscoe and the reviewers for providing useful feedback.

\bibliography{acl2016}
\bibliographystyle{acl2016}

\end{document}